\documentclass[12pt]{article}
\date{}
\usepackage{graphicx}
\usepackage[all]{xy}

\begin{document}
\title{Handling controversial arguments by matrix}
\author{{X\small{u} \large{Y}\small{uming}} \thanks{\footnotesize Corresponding author. E-mail: xuyuming@sdu.edu.cn } \\
{\footnotesize $School \ of \ Mathematics, \ Shandong \ University, \ Jinan, 250100, \ China$}
 }
\maketitle

\hspace{5mm} \begin{minipage}{12.5cm}

\begin{center}{\small \bf Abstract} \end{center}

{\small We introduce matrix and its block to the Dung's theory of argumentation framework. It is showed that each argumentation framework has a matrix representation, and the indirect attack relation and indirect defence relation can be determined by computing the matrix. This provide a powerful  mathematics way to find out the "controversial arguments" and deal with them in an argumentation framework. Also, we introduce several kinds of blocks based on the matrix of argumentation frameworks, and various prudent semantics of argumentation frameworks can be determined by computing and checking the matrices and their blocks which we have defined. In contrast with traditional method of reasoning and directed graph, our matric method has excellent advantages: computability (even can be realized on computer easily), entirety (all the needed "controversial arguments" or prudent extensions can be find out by checking the matrices and blocks obtained by computing). So, there is an intensive perspective to import the theory of matrices to the research of argumentation frameworks and its related areas. }

\par {\small \it  \hspace{0cm}Keywords:}
{\small Argumentation framework; controversial, prudent semantics; matrix; block}

\end{minipage} \\  \\

\hspace{-10mm} {\bf \large 1. Introduction}

\vspace{5mm}
In recent years, the area of argumentation begins to become increasingly central as a core study within Artificial Intelligence. A number of papers investigated and compared the properties of different semantics which have been proposed for argumentation frameworks (AFs, for short) as introduced by Dung [10, 4, 3, 11, 9, 19]. In early time, many of the analysis of arguments are expressed in natural language. Later on, a tradition of using diagrams has been developed to explicate the relations between the components of the arguments. Now, argumentation frameworks are usually represented as directed graphs, which play a significant role in modeling and analyzing the extension-based semantics of AFs. For further notations and techniques of argumentation, we refer the reader to [10, 18, 2, 17, 1].

This paper is a continuous work of [20]. Our aim is to introduce matrix as a new mathematic tool to the research of argumentation frameworks.
First, we assign a matrix of order $n$ for each argumentation framework with $n$ arguments. Each element of the matrix has only two possible values: one and zero, where one represents the attack relation and zero represents the non-attack relation between two arguments (they can be the same one). Under this circumstance, the matrix can be thought to be a representation of the argumentation framework. Second, we give the matrix methods to determine the indirect attack relation and indirect defence relation between two arguments. By this way, we can find the "controversial arguments" in an AF by computing the matrices of it.
Finally, we analysis the internal structure of the matrix corresponding to various prudent semantics of the argumentation framework, define several blocks corresponding to various prudent semantics, and give the matrix approaches to determine the stable p-extension, admissible p-extension and complete p-extension, which can be easily realized on computer.

As will be seen in later, the matrix of an argumentation framework is not only visualized as the directed graph, but also has another significant advantage on the aspect of computation. We shall study various prudent semantics of the argumentation framework by comparing and computing the matrix of the AF and its blocks. \\

\hspace{-10mm} {\bf \Large 2. Dung's theory of argumentation}

\vspace{5mm} Argumentation is a general approach to model defeasible reasoning and justification in Artificial Intelligence. So far, many theories of argumentation have been established [5, 7, 8]. Among them, Dung's theory of argumentation framework is quite influence. In fact, it is abstract enough to manage without any assumption on the nature of arguments and the attack relation between arguments. Let us first recall some basic notion in Dung's theory of argumentation framework. We restrict them to finite argumentation frameworks.

An argumentation framework is a pair $F = (A, R)$, where $A$ is a finite set of arguments and $R \subset A \times A$ represents the attack-relation. For $S \subset A$, we say that

(1) $S$ is conflict-free in $(A, R)$ if there are no $a, b \in S$ such that $(a, b) \in R$;

(2) $a \in A$ is defeated by $S$ in $(A, R)$ if there is $b \in S$ such that $(b, a) \in R$;

(3) $a \in A$ is defended by $S$ in $(A, R)$ if for each $b \in A$ with $(b, a) \in R$,

\hspace{5mm} we have $b$ is defeated by $S$ in $(A, R)$.

(4) $a \in A$ is acceptable with respect to $S$ if for each $b \in A$ with $(b, a) \in R$,

\hspace{5mm} there is some $c \in S$ such that $(c, b) \in R$.

\hspace{-10mm} Remark: For convenience, when $S = \{b\}$ has only one element, we will also say that $a$ is defended by $b$ instead of $S$.

The conflict-freeness, as observed by Baroni and Giacomin[1] in their study of evaluative criteria for extension-based semantics, is viewed as a minimal requirement to be satisfied within any computationally sensible notion of "collection of justified arguments". However, it is too weak a condition to be applied as a reasonable guarantor that a set of arguments is "collectively acceptable".

Semantics for argumentation frameworks can be given by a function $\sigma$ which assigns each AF $F = (A, R)$ a collection $\mathcal{S} \subset 2^{A}$ of extensions. Here, we mainly focus on the semantic $\sigma \in \{s, a, p, c, g, i, ss, e\}$ for stable, admissible, preferred, complete, grounded, ideal, semi-stable and eager extensions, respectively.

\hspace{-10mm} \textbf{Definition 1}[17] Let $F = (A, R)$ be an argumentation framework and $S \in A$.

(1) $S$ is a stable extension of $F$, $i.e.$, $S \in s(F)$, if $S$ is conflict-free in $F$ and each
$a \in A \setminus S$ is defeated by $S$ in $F$.

(2) $S$ is an admissible extension of $F$,  $i.e.$, $S \in a(F)$, if $S$ is conflict-free in $F$ and each
$a \in A \setminus S$ is defended by $S$ in $F$.

(3) $S$ is a preferred extension of $F$,  $i.e.$, $S \in p(F)$, if $S \in a(F)$ and for each $T \in a(F)$,
we have $S \not\subset T$.

(4)  $S$ is a complete extension of $F$,  $i.e.$, $S \in c(F)$, if $S \in a(F)$ and for each $a \in A$
defended by $S$ in $F$, we have $a \in S$.

(5)  $S$ is a grounded extension of $F$,  $i.e.$, $S \in g(F)$, if $S \in c(F)$ and for each $T \in c(F)$,
we have $T \not\subset S$.

(6)  $S$ is an ideal extension of $F$,  $i.e.$, $S \in i(F)$, if $S \in a(F)$, $S \subset \cap \{T: T \in p(F)\}$ and
for each $U \in a(F)$ such that $U \subset \cap \{T: T \in p(F)\}$, we have $S \not\subset U$.

(7)  $S$ is a semi-stable extension of $F$,  $i.e.$, $S \in ss(F)$, if $S \in a(F)$ and for each $T \in a(F)$,
we have $R^{+}(S) \not\subset R^{+}(T)$, where $R^{+}(U) = \{U \cap \{b: (a, b) \in R, A \in U\}\}$.

(8)  $S$ is a eager extension of $F$,  $i.e.$, $S \in e(F)$, if $S \in c(F)$, $S \subset \cap \{T: T \in ss(F)\}$ and
for each $U \in a(F)$ such that $U \subset \cap \{T: T \in ss(F)\}$, we have $S \not\subset T$.

\vspace{2mm} Note that, there are some elementary properties for any argumentation framework $F = (A, R)$ and semantic $\sigma$. If $\sigma \in \{a, p, c, g\}$, then we have $\sigma(F) \neq \emptyset$. And if $\sigma \in \{g, i, e\}$, then $\sigma(F)$ contains exactly one extension. Furthermore, the following relations hold for each argumentation framework $F = (A, R)$:

\vspace{2mm}\hspace{40mm}     $s(F) \subseteq p(F) \subseteq c(F) \subseteq a(F)$.

\vspace{2mm} Since every extension of an AF under the standard semantics (stable, preferred, complete and grounded extensions) introduced by Dung is an admissible set, the concept of admissible extensions plays an important role in the study of argumentation frameworks. \\

\hspace{-10mm} {\bf \Large 3.  Controversial arguments in framework}

\vspace{5mm} The controversial arguments was first defined when Dung discussed the coherence of argumentation frameworks in [13]. Then, Coste-Marquis, Devred and Marquis considered a refinement of the concept of "conflict-free set" in order to exclude "controversial arguments", i.e. arguments ${x, y}$ such that, although $(x, y) \notin R$ there is an "indirect attack" by $x$ on $y$: the resulting approach gives to the prudent semantics of [8]. Furthermore, Cayrol, Devred and Lagasquie-Schiex studied the "controversial arguments" in bipolar argumentation frameworks [5].

\hspace{-10mm} \textbf{Definition 2}[10] Let $F = (A, R)$ be an AF and $a, b \in A$.

\hspace{-10mm} (1) The argument $a$ indirectly attacks the argument $b$ iff there is an odd-length path from $a$ to $b$ in $F$, i.e., there is a finite sequence $a_{0}, a_{1}, ..., a_{2n+1}$ such that \ 1) $a = a_{0}$ and $b = a_{2n+1}$, and \ 2) for each $0 \leq i \leq 2n$, $a_{i}$ attacks $a_{i + 1}$.

\hspace{-10mm} (2) The argument $a$ indirectly defends the argument $b$ iff there is an even-length path from $a$ to $b$ in $F$ (length $\geq 2$), i.e., there is a finite sequence $a_{0}, a_{1}, ..., a_{2n}$ such that \ 1) $a = a_{0}$ and $b = a_{2n}$, and \ 2) for each $0 \leq i < 2n$, $a_{i}$ attacks $a_{i + 1}$.

\hspace{-10mm} (3) The argument $a$ is controversial $w.r.t$ the argument $b$ iff the argument $a$ indirectly attacks the argument $b$ and indirectly defends the argument $b$.

\hspace{-10mm} \textbf{Example 3} Let $F = (A, R)$, where $A = \{a, b, c, d, e, f\}$ and $R = \{(b, a), (c, a), (d,c), (e, b), $

\hspace{-10mm} $(e, d), (f, c)\}$. The directed graph  of $F$ is depicted on the followings:

$$
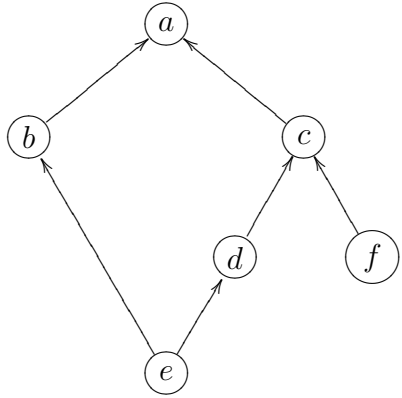
\xymatrix@C=5pt@W=15pt{ & & *++[o][F-]{a} & & &\\
                  *++[o][F-]{b}\ar[urr] & & & & *++[o][F]{c}\ar[ull] &\\
                  & & &*++[o][F-]{d}\ar[ur] & & *++[o][F-]{f}\ar[ul]\\
                  & &*++[o][F-]{e}\ar[uull]\ar[ur]& & &
}$$

It is easy to see that $a$ is attacked by $b$ and the defeater $e$ of $b$ is controversial $w.r.t$ $a$.
Even there is no direct conflict between $a$ and $e$, it seems uncautious  to accept  together both arguments in the same extension for the coherence of the extension set. This problem has some practical background in AI, and motivates the concept of prudent semantics and its study.

\hspace{-10mm} \textbf{Definition 4}[8] Let $F = (A, R)$ be an AF, and $S \in A$.

 (1) $S$ is p(rudent)-admissible iff every $a \in S$ is acceptable $w.r.t$ $S$ and $S$ is without indirect conflicts, i.e., there is no pair of arguments $a$ and $b$ of $S$ such that there is an odd-length path from $a$ to $b$ in $F$.

 (2) $S$ is a stable p-extension iff $S$ attacks every argument from $A \setminus S$ and without indirect conflicts.

 (3) $S$ is a preferred p-extension iff it is maximal $w.r.t$ $\subseteq$ among the p-admissible sets,

 (4) $S$ is a complete p-extension iff it is p-admissible, and every argument which is
 acceptable $w.r.t$ $S$ and without indirect conflicts with $S$ belongs to $S$.

  For prudent extensions of argumentation framework, there are several basic properties which can be easily deduced from the definition. On purpose of the self completeness, we list them in the following.

\hspace{-10mm} \textbf{Proposition 5}[8] Let $F = (A, R)$ be an AF, and $a, b \in A$.

(1) If $a$ is controversial $w.r.t$ $b$, then $\{a, b\}$ can not included into any p-admissible set.

(2) The set of all p-admissible subsets of $A$ is a complete set of $(2^{A}, \subseteq )$.

(3) For every p-admissible set $S \subseteq A$, there is at least one preferred p-extension $E \subseteq A$ such that $s \subseteq E$. \\

\hspace{-10mm} {\bf \Large 4. The matrix of argumentation frameworks}

\vspace{5mm} We know that the directed graph is a traditional tool in the research of argumentation framework, and has the feature of visualization. It is widely used for modeling and analyzing argumentation frameworks. In this paper, we will introduce the matrix representation of argumentation framework. Except for the visibility, it has a excellent advantage in analyzing argumentation framework and computing various semantics extension.

Let us first introduce some basic notation about matrix. An $m \times n$ matrix $A$ is a rectangular array of numbers, consisting of $m$ rows and $n$ columns, denoted by

\[
  A = \left(\begin{array}{cccccc}
  a_{1,1}&a_{1,2}&.&.&.&a_{1,n}\\
  a_{2,1}&a_{2,2}&.&.&.&a_{2,n}\\
  .&.&.&.&.&.\\
  a_{m,1}&a_{m,2}&.&.&.&a_{m,n}
  \end{array}\right).
\]
The $m \times n$ numbers $a_{1,1}, a_{1,2}, ..., a_{m,n}$ are the elements of the matrix $A$. We often called $a_{i,j}$ the $(i,j)$th element, and write $A = (a_{i, j})$. It is important to remember the first suffix of $a_{i,j}$ indicates the row and the second the column of $a_{i,j}$.

A column matrix is an $n \times 1$ matrix, and a row matrix is an $1 \times n$ matrix, denoted by

\[
  \left(\begin{array}{c}  x_{1}\\ x_{2}\\ .\\ .\\ .\\ x_{n}  \end{array}\right), \left(\begin{array}{cccccc}  x_{1}&x_{2}&.&.&.&x_{n}  \end{array}\right)
\]
respectively. Matrices of both these types can be regarded as vectors and referred to respectively as column vectors and row vectors.
Usually, the $i$-th row of a matrix $A$ is denoted by $A_{i, *}$, and the $j$-th column of $A$ is denoted by $A_{*, j}$.\\

\hspace{-10mm} \textbf{Definition 6} In a $m \times n$ matrix $A$ we specify any $k (\leq min\{m, n\})$ different rows $i_{1},i_{2},...,i_{k}$ and the same number of different columns. The elements appearing at the intersections of these rows and columns form a square matrix of order $k$. We call this matrix a principal block of order $k$ of the original matrix $A$; it is denoted by

\[
  M = \left(\begin{array}{cccccc}
  a_{i_{1}, i_{1}}&a_{i_{1}, i_{2}}&.&.&.&a_{i_{1}, i_{k}}\\
  a_{i_{2}, i_{1}}&a_{i_{2}, i_{2}}&.&.&.&a_{i_{2}, i_{k}}\\
  .&.&.&.&.&.\\
  a_{i_{k}, i_{1}}&a_{i_{k}, i_{2}}&.&.&.&a_{i_{k}, i_{k}}
  \end{array}\right),
\]
or $M = M_{i_{1},i_{2},...,i_{k}}^{i_{1},i_{2},...,i_{k}}$ for short, where $i_{1},i_{2},...,i_{k}$ are the numbers of the selected rows and columns.\\

\hspace{-10mm} \textbf{Definition 7} If in the original matrix $A$ we delete the rows and columns which make up the block $M$, then the remaining elements form a $(n-k) \times (m-k)$ matrix.
We call this matrix the complementary block of the block $M$, and is denoted by the symbol $\overline{M} = \overline{M_{i_{1},i_{2},...,i_{k}}^{i_{1},i_{2},...,i_{k}}}$.\\

\hspace{-10mm} \textbf{Definition 8} In a $m \times n$ matrix $A$ we specify any $k (\leq min\{m, n\})$ different rows $i_{1},i_{2},...,i_{k}$ and $h (\leq min\{m, n\})$ different columns $j_{1},j_{2},...,j_{h}$. The elements appearing at the intersections of these rows and columns form a $k \times h$ matrix. We call this matrix a block of order $k \times h$ of the original matrix $A$; it is denoted by

\[
  M = \left(\begin{array}{cccccc}
  a_{i_{1}, j_{1}}&a_{i_{1}, j_{2}}&.&.&.&a_{i_{1}, j_{h}}\\
  a_{i_{2}, j_{1}}&a_{i_{2}, j_{2}}&.&.&.&a_{i_{2}, j_{h}}\\
  .&.&.&.&.&.\\
  a_{i_{k}, j_{1}}&a_{i_{k}, j_{2}}&.&.&.&a_{i_{k}, j_{h}}
  \end{array}\right),
\]
or $M = M_{i_{1},i_{2},...,i_{k}}^{j_{1},j_{2},...,j_{h}}$ for short, where $i_{1},i_{2},...,i_{k}$ are the numbers of the selected rows and $j_{1},j_{2},...,j_{h}$ the numbers of the selected columns.

For the underlying set $A$ of the arguments of $AF$, we may enumerate it by using natural numbers. Contrasting with the form $A = \{a, b, ...\}$, it is more convenience to put $A = \{1, 2, ..., n\}$ if the cardinality of $A$ is large. In fact, this arrangement has an obvious advantage for computing, and we will follow this arrangement in the below discussion.

\vspace{2mm}
\hspace{-10mm} \textbf{Definition 9} Let $F = (A, R)$ be an $AF$, in which the cardinality of $A$ is $n$. The matrix of $F$ is an $n \times n$ matrix, its entries is determined by the following rule:

(1) $a_{i,j} = 1$ if $(i, j) \in R$;

(2) $a_{i,j} = 0$ if $(i, j) \notin R$;

\vspace{2mm}
\hspace{-10mm} \textbf{Example 10} Consider the argumentation framework $F = (A, R)$, where $A = \{1,2,3\}$ and $R = \{(1, 2), (2, 3), (3, 1)\}$. By the definition we have the following matrix of $F = (A, R)$:

\[
  \left(\begin{array}{ccc}
  0&1&0\\
  0&0&1\\
  1&0&0
  \end{array}\right)
\]

 \vspace{2mm}
\hspace{-10mm} \textbf{Example 11} Given an argumentation framework $F = (A, R)$, where $A = \{1,2,3,4\}$ and $R = \{(1, 2), (1, 3), (2, 1), (2, 3), (3, 4)\}$. By the definition we have the following matrix of $F = (A, R)$:

\[
  \left(\begin{array}{cccc}
  0&1&1&0\\
  1&0&1&0\\
  0&0&0&1\\
  0&0&0&0
  \end{array}\right)
\]

In comparison with graph-theoretic way and mathematical logic way, the matrix of an argumentation framework has many excellent features. First, it possess a concise mathematical format. Secondly, it contains all information of the $AF$ by combining the arguments with attack relation in a specific manner in the matrix $M(F)$. Also, it can be deal with by program on computer. The most important is that we can import the knowledge of matrix to the research of argumentation frameworks. \\

\hspace{-10mm} {\bf \Large 5. Handling controversial arguments by matrix in AF}

\vspace{5mm}
\hspace{-10mm} \textbf{Example 12} Given an argumentation framework $F = (A, R)$, where $A = \{1,2,3,4,5,6\}$ and $R = \{(2, 1), (3, 1), (4, 3), (5, 2), (5, 4), (6,3)\}$. Then, $F$ can be represented by a directed graph as follows.

$$
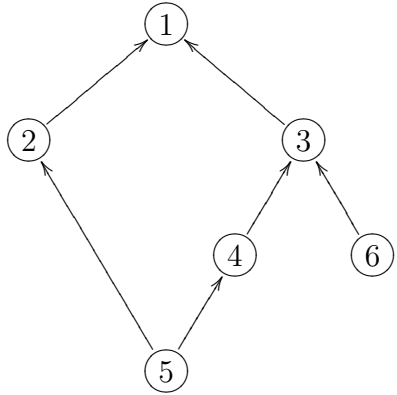
\xymatrix@C=5pt@W=15pt{ & & *++[o][F-]{1} & & &\\
                  *++[o][F-]{2}\ar[urr] & & & & *++[o][F]{3}\ar[ull] &\\
                  & & &*++[o][F-]{4}\ar[ur] & & *++[o][F-]{6}\ar[ul]\\
                  & &*++[o][F-]{5}\ar[uull]\ar[ur]& & &
}$$

It is obviously that $1$ is defended by $5$. Next, we study the structure of the matrix of $F = (A, R)$ to find the reflection of defence relation between $5$ and $1$. First, let us write out the matrix of $F$;

\[
  \left(\begin{array}{cccccc}
  0&0&0&0&0&0\\
  1&0&0&0&0&0\\
  1&0&0&0&0&0\\
  0&0&1&0&0&0\\
  0&1&0&1&0&0\\
  0&0&1&0&0&0
  \end{array}\right).
\]

In the column vector $F_{*, 1}$(column 1), $a_{2, 1} = 1$ means that $(2, 1) \in R$, and thus the argument $2$ attacks the argument $1$. In the row vector $F_{5, *}$(row 5), $a_{5, 2} = 1$ means that $(5, 2) \in R$, and thus the argument $5$ attacks the argument $2$. By combination,  $a_{2, 1} = 1$ and $a_{5, 2} = 1$ in the matrix $M(F)$ ensure the fact that $1$ is defended by $5$. This can be come down to the element $b_{5, 1} \neq 0$ in the matrix $A^{2} = B = (b_{i, j})$. By a similar discussion, we can see that $a_{3, 1} = 1$ and $a_{4, 3} = 1$ in the matrix $M(F)$ play the similar role to guarantee that $1$ is defended by $4$.

In the column vector $F_{*, 3}$(column 3), $a_{4, 3} = 1$ means that $(4, 3) \in R$, and thus the argument $4$ attacks the argument $3$. In the row vector $F_{*, 5}$(row 5), $a_{5, 4} = 1$ means that $(5, 4) \in R$, and thus the argument $5$ attacks the argument $4$. Therefore, in the matrix $M(F)$ $a_{4, 3} = 1$ and $a_{5, 4} = 1$ ensure the fact that $3$ is defended by $5$. This can be come down to the element $b_{5, 3} \neq 0$ in the matrix $A^{2} = B = (b_{i, j})$.

Further analysis indicates that the converse is also true. And, we can generalize this idea to obtain a matric method, by which the defence relation between two arguments will be easily be determined.

\vspace{2mm}
\hspace{-10mm} \textbf{Theorem 13} Given an argumentation framework $F = (A, R)$ with $A = \{1, 2, ..., n\}$, $i, j \in A (1 \leq i, j \leq n)$. Then, $j$ is defended by $i$ in $F$ iff $b_{i, j} \neq 0$ in the matrix $M^{2}(F) = B = (b_{i, j})$.

\vspace{2mm}
\hspace{-10mm} \textbf{Proof} Assume that  $j$ is defended by $i$ in $F$, then there is some $1 \leq t \leq n$ such that the argument $i$ attacks the argument $t$, and the argument $t$ attacks the argument $j$. This implies that $(i, t) \in R$ and $(t, j) \in R$. Therefore, we have that $a_{i, t} = 1$ and $a_{t, j} = 1$. Since $a_{i, t}$ is at the intersection of row $i$ and column $t$, $a_{t, j}$ is at the intersection of row $t$ and column $j$, we have that $b_{i, j} = a_{i, 1}a_{1, j} + ... + a_{i, t}a_{t, j} + ... + a_{i, n}a_{n, j} \neq 0$ in the matrix $M^{2}(F) = B$.

Conversely, suppose that $b_{i, j} \neq 0$ in the matrix $M^{2}(F) = B$, i.e., $b_{i, j} = a_{i, 1}a_{1, j} + ... + a_{i, t}a_{t, j} + ... + a_{i, n}a_{n, j} \neq 0$. Then, there is some $1 \leq t \leq n$ such that $a_{i, t}a_{t, j} \neq 1$. This implies that $a_{i, t} = 1$ and $a_{t, j} = 1$, i.e., $(i, t) \in R$ and $(t, j) \in R$. It follows that the argument $i$ attacks the argument $t$, and the argument $t$ attacks the argument $j$. This induce that the argument $j$ is defended by the argument $i$.

\hspace{-10mm} Remark: From the proof of the above theorem, we can deduce that $b_{i, j} = a_{i, 1}a_{1, j} + ... + a_{i, t}a_{t, j} + ... + a_{i, n}a_{n, j} = k$ if and only if, there are $k$ different paths from $i$ to $j$ in $F$, whose length is 2.\\

\hspace{-10mm} \textbf{Example (cont)} In this example, we can also find out that the argument $5$ indirectly attacks the argument $1$, while the argument $1$ is defended by the argument $5$. Since the argument $1$ is defended by the argument $4$, and the argument $5$ attacks the argument $4$.

Next, we study the structure of the matrix $M(F)$ of $F$ to find the reflection of indirect attack relation between $5$ and $1$.

In the column vector $M^{2}(F)_{*, 1}$(column 1), $b_{4, 1} = 1$ means that the argument $1$ is defended by the argument $4$. In the row vector $F_{5, *}$(row 5), $a_{5, 4} = 1$ means that $(5, 4) \in R$, i.e., the argument $5$ attacks the argument $4$. By combination, $b_{4, 1} = 1$ in the matrix $M^{2}(F) = B$ and $a_{5, 4} = 1$ in the matrix $M(F)$ ensure the fact that the argument $5$ indirectly attacks the argument $1$. This can be come down to the element $c_{5, 1} \neq 0$ in the matrix $M^{3}(F) = C = (c_{i, j})$.

Further discussion tell us that the converse is also true. We generalize this idea and give the matric method to determine the indirect attack relation between two arguments as follows.

\vspace{2mm}
\hspace{-10mm} \textbf{Theorem 14} Given an argumentation framework $F = (A, R)$ with $A = \{1, 2, ..., n\}$, $i, j \in A (1 \leq i, j \leq n)$. Then, $i$ indirectly attacks $j$ in $F$ iff there is some odd number $k$ such that $m(k)_{i, j} \neq 0$ in the matrix $M^{k}(F) = M(k) = (m(k)_{s, t})$.

\vspace{2mm}
\hspace{-10mm} \textbf{Proof} Assume that  $i$ indirectly attacks $j$ in $F$, then there are  $1 \leq i_{1}, i_{2},...,i_{k-1} \leq n$ such that the argument $i$ attacks the argument $i_{1}$, the argument $i_{1}$ attacks the argument $i_{2}$, ..., and the argument $i_{k-1}$ attacks the argument $j$, where $k$ is an odd number. This implies that $(i, i_{1}), (i_{1}, i_{2}), ..., (i_{k-1}, j) \in R$. Therefore, we have that $a_{i, i_{1}} = 1, a_{i_{1}, i_{2}} = 1, ..., a_{i_{k-1}, j} = 1$. Since $a_{i, i_{1}}$ is at the intersection of row $i$ and column $i_{1}$, $a_{i_{1}, i_{2}}$ at the intersection of row $i_{1}$ and column $i_{2}$, ..., $a_{i_{k-1}, j}$ at the intersection of row $i_{k-1}$ and column $j$, we have that

$m(2)_{i, i_{2}} = a_{i, 1}a_{1, i_{2}} + ... + a_{i, i_{1}}a_{i_{1}, i_{2}} + ... + a_{i, n}a_{n, i_{2}} \neq 0$ in the matrix $M^{2}(F) = M(2) = (m(2)_{s, t})$, 

$m(3)_{i, i_{3}} = m(2)_{i, 1}a_{1, i_{3}} + ... + m(2)_{i, i_{2}}a_{i_{2}, i_{3}} + ... + m(3)_{i, n}a_{n, i_{3}} \neq 0$ in the matrix 
$M^{3}(F) = M^{2}(F)M(F) = M(3) = (m(3)_{s, t})$,

 ......................................................................................,

$m(k)_{i, j} = m(k-1)_{i, 1}a_{1, j} + ... + m(k-1)_{i, i_{k-1}}a_{i_{k-1}, j} + ... + m(k-1)_{i, n}a_{n, j} \neq 0$ in the 
matrix $M^{k}(F) = M^{k-1}(F)M(F) = M(k) = (m(k)_{s, t})$.

Conversely, suppose that $m(k)_{i, j} \neq 0$ in the matrix $M^{k}(F) = M^{k-1}(F)M(F)$, i.e., $m(k-1)_{i, 1}a_{1, j} + m(k-1)_{i, 2}a_{2, j} + ... + m(k-1)_{i, n}a_{n, j} \neq 0$, where $k$ is an odd number. Then, there is some $1 \leq i_{1} \leq n$ such that $a_{i_{1}, j} = 1$ (in the matrix $M(F) = (a_{i, j})$) and $m(k-1)_{i, i_{1}} \neq 0$ (in the matrix $M^{k-1}(F) = (m(k-1)_{s, t})$). Since $m(k-1)_{i, i_{1}} = m(k-2)_{i, 1} a_{1, i_{1}} + m(k-2)_{i, 2} a_{2, i_{1}} + ... + m(k-2)_{i, n} a_{n, i_{1}} \neq 0$ in the matrix $M^{k-1}(F) = M^{k-2}(F)M(F)$, there is some $1 \leq i_{2} \leq n$ such that $a_{i_{2}, i_{1}} = 1$ (in the matrix $M(F)$) and $m(k-2)_{i, i_{2}} \neq 0$ (in the matrix $M^{k-2}(F) = (m(k-2)_{s, t})$). By similar discussion, we can find out $1 \leq i_{3}, i_{4}, ..., i_{k-1} \leq n$ such that $a_{i_{3}, i_{2}} = 1, a_{i_{4}, i_{3}} = 1, ..., a_{i_{k-1}, i_{k-2}} = 1, a_{i, i_{k-1}} = 1$. Therefore, we have $(i, i_{k-1}), (i_{k-1}, i_{k-2}), ..., (i_{3}, i_{2}), (i_{2}, i_{1}), (i_{1}, j) \in R$, i.e., the argument $i$ attacks the argument $i_{k-1}$, the argument $i_{k-1}$ attacks the argument $i_{k-2}$, ..., the argument $i_{1}$ attacks the argument $j$. Since $k$ is an odd number, we conclude that the argument $i$ indirectly attacks the argument $j$.

For the indirect defence relation between two arguments in an argumentation framework, we have a similar result as follows, which can be  proved by referring to the proof of theorem 13 and theorem 14.

\vspace{2mm}
\hspace{-10mm} \textbf{Theorem 15} Given an argumentation framework $F = (A, R)$ with $A = \{1, 2, ..., n\}$, $i, j \in A (1 \leq i, j \leq n)$. Then, $i$ indirectly defends $j$ in $F$ iff there is some even number $k$ which is greater than 2 such that $m(k)_{i, j} \neq 0$ in the matrix $M^{k}(F) = (m(k)_{s, t})$.

\hspace{-10mm} Remark: In theorem 15, the determination of indirect defence relation between two arguments does not involve $k =2$, but it is obvious that theorem 13 and theorem 15 does have the same core feature. For the sake of coherence, we rewrite theorem 13 as follows.

\vspace{2mm}
\hspace{-10mm} \textbf{Theorem 16} Given an argumentation framework $F = (A, R)$ with $A = \{1, 2, ..., n\}$, $i, j \in A (1 \leq i, j \leq n)$. Then, $i$ defends $j$ in $F$ iff  $m(2)_{i, j} \neq 0$ in the matrix $M^{2}(F) = (m(2)_{s, t})$.

From the above discussion, we summarize a matrix method for determining the "controversial arguments" in AF. It is important in the theoretic sense, and we will refine it to a very perfect grade later on.

\vspace{2mm}
\hspace{-10mm} \textbf{Theorem 17} Given an argumentation framework $F = (A, R)$ with $A = \{1, 2, ..., n\}$, $i, j \in A (1 \leq i, j \leq n)$. Then, $i$ is controversial $w.r.t$ $j$ iff there are odd number $3 \leq k \leq n$ and even number $2 \leq l \leq n$ such that $m(k)_{i, j} \neq 0$ in the matrix $M^{k}(F) = M(k)= (m(k)_{s, t})$ and $m(l)_{i, j} \neq 0$ in the matrix $M^{l}(F) = M(l)= (m(l)_{s, t})$.

Further observation indicates that the odd number $k$ in theorem 14 and the even number $k$ in theorem 15 can be limited to not greater than $n$. This fact will bring us a great deal of benefit in determining the indirect attack relation and indirect defence relation between two arguments in AF by computing the matrix $M^{k}(F)$.

\vspace{2mm}
\hspace{-10mm} \textbf{Theorem 18} Given an argumentation framework $F = (A, R)$ with $A = \{1, 2, ..., n\}$. If there is a path from $i_{1}$ to $i_{k+1}$ satisfying that $(i_{1}, i_{2}), (i_{2}, i_{3}), ..., (i_{k}, i_{k+1}) \in R$. Then, there is some number $r \leq n$ such that $m(i_{1}, i_{k+1}) \neq 0$ in the matrix $M^{m}(F)$.

\vspace{2mm}
\hspace{-10mm} \textbf{Proof} Without lost of generality, we assume that $i_{1} \leq i_{k+1}$ and $i_{t} \neq i_{t+1}$ for $1 \leq t \leq k$. For the argument $i_{2} \in A$, we put the $t_{1}$ to be the last one in the sequence $2, 3, ..., k$ (correspond to the sequence $i_{2}, i_{3}, ..., i_{k}$) such that $i_{t_{1}}$ equals to $i_{2}$. Then, there is a path from $i_{1}$ to $i_{k+1}$ satisfying that $(i_{1}, i_{t_{1}}), (i_{t_{1}}, i_{t_{1}+1}), (i_{t_{1}+1}, i_{t_{1}+2}), ..., (i_{k}, i_{k+1}) \in R$.  For the argument $i_{t_{1}+1}$, we put the $t_{2}$ to be the last one in the sequence $t_{1} + 1, t_{1} + 2, ..., k$ such that $i_{t_{2}}$ equals to $i_{t_{1} + 1}$. Then, there is a path from $i_{1}$ to $i_{k+1}$ satisfying that

$(i_{1}, i_{t_{1}}), (i_{t_{1}}, i_{t_{2}}), (i_{t_{2}}, i_{t_{2}+1}), ..., (i_{k}, i_{k+1}) \in R$.

\hspace{-10mm} This process will be end at some step, and we finally obtain a path from $i_{1}$ to $i_{k+1}$ such that
$(i_{1}, i_{t_{1}}), (i_{t_{1}}, i_{t_{2}}), (i_{t_{2}}, i_{t_{3}}), ..., (i_{t_{r-1}}, i_{t_{r}}), (i_{t_{r}}, i_{k+1}) \in R$. It follows that $a_{i_{1}, i_{t_{1}}} = 1, a_{i_{t_{1}}, i_{t_{2}}} = 1, ..., a_{i_{t_{r}}, i_{k+1}} = 1$, and thus we have $m(r)_{i_{1}, i_{k+1}} \neq 0$ in the matrix $M^{r}(F)$ (just as the proof in first paragraph of theorem 14). From the selection of $i_{t_{1}}, i_{t_{2}}, ..., i_{t_{r}}$, we conclude that they are different from each other, and thus $r \leq n$.

\vspace{2mm} For each argumentation framework $F = (A, R)$ with $A = \{1, 2, ..., n\}$. If $n = 2k$ is an even number, then we define the matrix $M^{E}(F) = (m(E)_{s, t}) = M^{2}(F) + M^{4}(F) + ..., M^{2k}(F)$, and $M^{O}(F) = (m(O)_{s, t}) = M^{3}(F) + M^{5}(F) + ..., M^{2k - 1}(F)$. If $n = 2k + 1$ is an odd number, then we define the matrix $M^{E}(F) = (m(E)_{s, t}) = M^{2}(F) + M^{4}(F) + ... , M^{2k}(F)$, and $M^{O}(F) = (m(O)_{s, t}) = M^{3}(F) + M^{5}(F) + ..., M^{2k + 1}(F)$.

By theorem 14, theorem 15, and theorem 16, we claim that $m(E)_{i, j} \neq 0$ in the matrix $M^{E}(F)$ if and only if there is some $k \geq 1$ such that $m(2k)_{i, j} \neq 0$ in the matrix $M^{2k}(F) = M(2k) = (m(2k)_{s,t})$, and $m(O)_{i, j} \neq 0$ in the matrix $M^{O}(F)$ if and only if there is some $k \geq 1$ such that $m(2k+1)_{i, j} \neq 0$ in the matrix $M^{2K+1}(F)  = M(2k+1) = (m(2k+1)_{s,t})$. With this preparation, we can improve the above matrix method to be a powerful tool for finding out the "controversial arguments" in AF as follows.

\vspace{2mm}
\hspace{-10mm} \textbf{Corollary 19} Given an argumentation framework $F = (A, R)$ with $A = \{1, 2, ..., n\}$, $i, j \in A (1 \leq i, j \leq n)$. Then, $i$ is controversial $w.r.t$ $j$ iff $m(O)_{i, j} \neq 0$ in the matrix $M^{O}(F) = (m(O)_{s, t})$ and $m(E)_{i, j} \neq 0$ in the matrix $M^{E}(F) = (m(E)_{s, t})$.

\vspace{2mm} By this corollary, we only need to computer the two matrices $M^{O}(F)$ and $M^{E}(F)$ for any argumentation framework $F = (A, R)$ with $A = \{1, 2, ..., n\}$. Then, by checking the elements of them we will find all the "controversial arguments" in $F$. If both $m(O)_{i, j}$ and $m(E)_{i, j}$ are nonzero, then $i$ is controversial $w.r.t$ $j$. Otherwise, $i$ is not controversial $w.r.t$ $j$. Compare with checking directed graph, computing matrices has great advantage in determining the "controversial arguments". We only need to computer the matrices without any comparing and reasoning on the directed graph. Especially when the AF has large number of arguments, writing a directed graph is not an easy thing. Furthermore, the computing of matrices can be carry out on computer easily.   \\

\hspace{-10mm} {\bf \Large 6. Determination of the stable p-extensions}

For convenience, from this section we assume that the sequences $i_{1}, i_{2}, ..., i_{k}$ and $j_{1}, j_{2}, ..., j_{h}$ are all increasing.

\vspace{2mm}
\hspace{-10mm} \textbf{Definition 20}[20] Let $F = (A, R)$ be an argumentation framework with $A = \{1, 2, ..., n\}$, $M(F) = (a_{i, j})$ is the matrix of $F$, and $S = \{i_{1}, i_{2}, ..., i_{k}\} \subset A$. The principal block

\[
  M^{i_{1}, i_{2}, ..., i_{k}}_{i_{1}, i_{2}, ..., i_{k}} = \left(\begin{array}{cccccc}
  a_{i_{1}, i_{1}}&a_{i_{1}, i_{2}}&.&.&.&a_{i_{1}, i_{k}}\\
  a_{i_{2}, i_{1}}&a_{i_{2}, i_{2}}&.&.&.&a_{i_{2}, i_{k}}\\
  .&.&.&.&.&.\\
  a_{i_{k}, i_{1}}&a_{i_{k}, i_{2}}&.&.&.&a_{i_{k}, i_{k}}
  \end{array}\right)
\]
of order $k$ in the matrix $M(F)$ is called the $cf$-block of $S$, and denoted by $M^{cf}$ for short.

In other words, the elements appearing at the intersection of rows $i_{1}, i_{2}, ..., i_{k}$ and the same number columns in the matrix $M(F)$ form the $cf$-block $ M^{cf}$ of $S$.

\vspace{2mm}
\hspace{-10mm} \textbf{Definition 21}[20] Let $F = (A, R)$ be an argumentation framework with $A = 1, 2, ..., n$, $M(F) = (a_{i, j})$ is the matrix of $F$, and $S = \{i_{1}, i_{2}, ..., i_{k}\} \subset A$ is a stable extension of $F$. We say that the $k \times h$ block

\[
  M^{i_{1}, i_{2}, ..., i_{k}}_{j_{1}, j_{2}, ..., j_{h}} = \left(\begin{array}{cccccc}
  a_{i_{1}, j_{1}}&a_{i_{1}, j_{2}}&.&.&.&a_{i_{1}, j_{h}}\\
  a_{i_{2}, j_{1}}&a_{i_{2}, j_{2}}&.&.&.&a_{i_{2}, j_{h}}\\
  .&.&.&.&.&.\\
  a_{i_{k}, j_{1}}&a_{i_{k}, j_{2}}&.&.&.&a_{i_{k}, j_{h}}
  \end{array}\right)
\]
in the matrix $M(F)$ is the $s$-block of $S$, where $\{j_{1}, j_{2}, ..., j_{h}\} = A \setminus S$. Usually, we denote it by $M^{s}$ for short.

Namely, the elements appearing at the intersection of rows $i_{1}, i_{2}, ..., i_{k}$ and columns $j_{1}, j_{2}, ..., j_{h}$ in the matrix $M(F)$ form the $s$-block $M^{s}$ of $S$.

\vspace{2mm} For any argumentation framework $F = (A, R)$ with $A = \{1, 2, ..., n\}$, let $M(F) = (a_{i, j})$ be its matrix. We consider the matrix  $M^{O}(F) = (m(O)_{i, j}) = M^{3}(F) + M^{5}(F) + ...+ M^{2k+1}(F)$, where $2k+1$ is the greatest natural number which is less than $n$. It is obvious that $m(O)_{i, j} = m(3)_{i, j} + m(5)_{i, j} + ... + m(2k+1)_{i, j}$. By theorem 14 and corollary 19, the argument $i$ indirectly attacks the argument $j$ if and only if there is some natural number $1 \leq t \leq k$ such that  $m(2t+1)_{i, j} = 1$ which is equivalent to $m(O)_{i, j} \neq 0$. This fact leads to the following definition.

\vspace{2mm}
\hspace{-10mm} \textbf{Definition 22} Let $F = (A, R)$ be an argumentation framework with $A = \{1, 2, ..., n\}$, and $S = \{i_{1}, i_{2}, ..., i_{k}\} \subset A$ is a stable p-extension of $F$. We say that the block

\[
  M_{p}(O)^{i_{1}, i_{2}, ..., i_{k}}_{i_{1}, i_{2}, ..., i_{k}} = \left(\begin{array}{cccccc}
  m(O)_{i_{1}, i_{1}}&m(O)_{i_{1}, i_{2}}&.&.&.&m(O)_{i_{1}, i_{k}}\\
  m(O)_{i_{2}, i_{1}}&m(O)_{i_{2}, i_{2}}&.&.&.&m(O)_{i_{2}, i_{k}}\\
  .&.&.&.&.&.\\
  m(O)_{i_{k}, i_{1}}&m(O)_{i_{k}, i_{2}}&.&.&.&m(O)_{i_{k}, i_{k}}
  \end{array}\right),
\]
of order $k$ in the matrix $M^{O}(F)$ of $F$ is the $p$-block of $S$.

In fact, the elements appearing at the intersection of rows $i_{1}, i_{2}, ..., i_{k}$ and the same number columns in the matrix $M^{O}(F)$ form the $p$-block $M_{p}(O)^{i_{1}, i_{2}, ..., i_{k}}_{i_{1}, i_{2}, ..., i_{k}}$ of $S$.

\vspace{2mm}
\hspace{-10mm} \textbf{Theorem 23}[20] Given an argumentation framework $F = (A, R)$ with $A = \{1, 2, ..., n\}$, then $S = \{i_{1}, i_{2}, ..., i_{k}\}  \subset A$ is a stable extension in $F$ iff the $cf$-block $M^{cf} = M^{i_{1}, i_{2}, ..., i_{k}}_{i_{1}, i_{2}, ..., i_{k}}$ of $S$ is zero and every column vector of its $s$-block $M^{s} = M^{i_{1}, i_{2}, ..., i_{k}}_{j_{1}, j_{2}, ..., j_{h}}$ are non-zero, where $A \setminus S = \{j_{1}, j_{2}, ..., j_{h}\}$.

\vspace{2mm}
\hspace{-10mm} \textbf{Theorem 24} Given an argumentation framework $F = (A, R)$ with $A = \{1, 2, ..., n\}$, then $S = \{i_{1}, i_{2}, ..., i_{k}\}  \subset A$ is a stable p-extension in $F$ iff the following conditions hold:

(1) The $cf$-block $M^{cf} = M^{i_{1}, i_{2}, ..., i_{k}}_{i_{1}, i_{2}, ..., i_{k}}$ of $S$ is zero,

(2) Every column vector of the $s$-block $M^{s}$ of $S$ is non-zero, where $A \setminus S = \{j_{1}, j_{2}, ..., j_{h}\}$,

(3) The p-block $M_{p}(O)^{i_{1}, i_{2}, ..., i_{k}}_{i_{1}, i_{2}, ..., i_{k}}$ of $S$ is zero.

\vspace{2mm}
\hspace{-10mm} \textbf{Proof} Assume that $S$ is a stable p-extension, then the conditions (1) and (2) hold by definition 4 and theorem 23. For any $i_{s}, i_{t} \in S (1 \leq s, t \leq k)$, we know that the argument $i_{t}$ does not indirectly attack the argument $i_{s}$. So, we have that $m(O)_{i_{t}, i_{s}} = 0$ by corollary 19. It follows that the p-block $M_{p}(O)^{i_{1}, i_{2}, ..., i_{k}}_{i_{1}, i_{2}, ..., i_{k}}$ of $S$ is zero.

Conversely, suppose that the conditions (1), (2) and (3) hold. Then, by theorem 23 $S$ is firstly a stable extension. For any $i_{t}, i_{s} \in S (1 \leq s, t \leq k)$, we have $m(O)_{i_{t}, i_{s}} = 0$ by condition (3). It follows that the argument $i_{t}$ does not indirectly attack the argument $i_{s}$. Therefore, we conclude that $S$ is a stable p-extension.

\hspace{-10mm} Remark: In this theorem, condition (1) ensures that $S$ is a conflict-free set. And, condition (2) shows the feature of $S$ that there are no indirect attack relation between any two arguments in $S$. \\

\hspace{-10mm} {\bf \Large 7. Determination of the admissible p-extensions}

\vspace{2mm}
\hspace{-10mm} \textbf{Definition 25}[20] Let $F = (A, R)$ be an argumentation framework with $A = \{1, 2, ..., n\}$, and $S = \{i_{1}, i_{2}, ..., i_{k}\} \subset A$ be an admissible extension of $F$. Then, the $h \times k$ block

\[
  M^{j_{1}, j_{2}, ..., j_{h}}_{i_{1}, i_{2}, ..., i_{k}} = \left(\begin{array}{cccccc}
  a_{j_{1}, i_{1}}&a_{j_{1}, i_{2}}&.&.&.&a_{j_{1}, i_{k}}\\
  a_{j_{2}, i_{1}}&a_{j_{2}, i_{2}}&.&.&.&a_{j_{2}, i_{k}}\\
  .&.&.&.&.&.\\
  a_{j_{h}, i_{1}}&a_{j_{h}, i_{2}}&.&.&.&a_{j_{h}, i_{k}}
  \end{array}\right)
\]
in the matrix $M(F)$ of $F$ is called the $a$-block of $S$, where $\{j_{1}, j_{2}, ..., j_{h}\} = A \setminus S$, and denoted by $M^{a}$ for short.

In other words, the elements appearing at the intersection of rows $j_{1}, j_{2}, ..., j_{h}$ and columns $i_{1}, i_{2}, ..., i_{k}$ in the matrix $M(F)$ of $F$ form the $a$-block $M^{a} = M^{j_{1}, j_{2}, ..., j_{h}}_{i_{1}, i_{2}, ..., i_{k}}$ of $S$.

\vspace{2mm}
\hspace{-10mm} \textbf{Theorem 26}[20] Given an argumentation framework $F = (A, R)$ with $A = \{1, 2, ..., n\}$, then $S = \{i_{1}, i_{2}, ..., i_{k}\} \subset A$ is an admissible extension in $F$ iff the following conditions hold:

(1) The $cf$-block $M^{cf} = M^{i_{1}, i_{2}, ..., i_{k}}_{i_{1}, i_{2}, ..., i_{k}}$ of $S$ is zero,

 (2) The column vector of $s$-block $M^{s}$ of $S$ corresponding to the non-zero row vector of
the $a$-block $M^{a}$ of $S$ is non-zero, where $A \setminus S = \{j_{1}, j_{2}, ..., j_{h}\}$.

According to the theorem 26 and the proof of theorem 24, it is easy to deduce the following result by which we can determine all admissible p-extension in any argumentation framework.

\vspace{2mm}
\hspace{-10mm} \textbf{Theorem 27} Given an argumentation framework $F = (A, R)$ with $A = \{1, 2, ..., n\}$, then $S = \{i_{1}, i_{2}, ..., i_{k}\} \subset A$ is an admissible p-extension in $F$ iff the following conditions hold:

(1) The $cf$-block $M^{cf} = M^{i_{1}, i_{2}, ..., i_{k}}_{i_{1}, i_{2}, ..., i_{k}}$ of $S$ is zero,

(2) The column vector of $s$-block $M^{s}$ of $S$ corresponding to the non-zero row vector of
the $a$-block $M^{a}$ of $S$ is non-zero, where $A \setminus S = \{j_{1}, j_{2}, ..., j_{h}\}$.

(3) The p-block $M_{p}(O)^{i_{1}, i_{2}, ..., i_{k}}_{i_{1}, i_{2}, ..., i_{k}}$ of $S$ is zero.

\vspace{2mm}
\hspace{-10mm} Remark: From the above theorem, we conclude that any stable p-extension must be a p-admissible set. And, this relation is clearly expressed by the $s$-block $M^{s} = M^{i_{1}, i_{2}, ..., i_{k}}_{j_{1}, j_{2}, ..., j_{h}}$ of $S$ in the matrix $M(F)$, i.e., the condition every column vector of its $s$-block $M^{s} = M^{i_{1}, i_{2}, ..., i_{k}}_{j_{1}, j_{2}, ..., j_{h}}$ are non-zero is stronger than that the column vector of $s$-block $M^{s} = M^{i_{1}, i_{2}, ..., i_{k}}_{j_{1}, j_{2}, ..., j_{h}}$ of $S$ corresponding to the non-zero row vector of the $a$-block $M^{a} = M^{j_{1}, j_{2}, ..., j_{h}}_{i_{1}, i_{2}, ..., i_{k}}$ of $S$ are non-zero.

For the determination of preferred p-extensions of an argumentation framework, we may compare all the p-admissible sets to find the maximal ones. By proposition 5, each maximal p-admissible set is a preferred p-extension of the AF.\\

\hspace{-10mm} {\bf \Large 8.  Determination of the complete p-extensions}

\vspace{2mm}
\hspace{-10mm} \textbf{Definition 28}[20] Let $F = (A, R)$ be an argumentation framework with $A = \{1, 2, ..., n\}$, and $S = \{i_{1}, i_{2}, ..., i_{k}\} \subset A$ is a complete extension of $F$. We say that the block

\[
  M^{j_{1}, j_{2}, ..., j_{h}}_{j_{1}, j_{2}, ..., j_{h}} = \left(\begin{array}{cccccc}
  a_{j_{1}, i_{1}}&a_{j_{1}, i_{2}}&.&.&.&a_{j_{1}, i_{k}}\\
  a_{j_{2}, i_{1}}&a_{j_{2}, i_{2}}&.&.&.&a_{j_{2}, i_{k}}\\
  .&.&.&.&.&.\\
  a_{j_{h}, i_{1}}&a_{j_{h}, i_{2}}&.&.&.&a_{j_{h}, i_{k}}
  \end{array}\right)
\]
of order $h$ in the matrix $M(F)$ of $F$ is the $c$-block of $S$, where $\{j_{1}, j_{2}, ..., j_{h}\} = A \setminus S$. We often write it $M^{c}$ for short.

In fact, the elements appearing at the intersection of rows $j_{1}, j_{2}, ..., j_{h}$ and the same number columns in the matrix $M(F)$ form the $c$-block $M^{c} = M^{j_{1}, j_{2}, ..., j_{h}}_{j_{1}, j_{2}, ..., j_{h}}$ of $S$.

\vspace{2mm}
\hspace{-10mm} \textbf{Lemma 29}[20] Let $F = (A, R)$ be an argumentation framework with $A = \{1, 2, ..., n\}$, then $S = \{i_{1}, i_{2}, ..., i_{k}\} \subset A$ is a complete extension of $F$ iff $S$ is an admissible set and each $j_{t} \in S (1 \leq t \leq h)$ is not defended by $S$.

\vspace{2mm}
\hspace{-10mm} \textbf{Theorem 30}[20] Given an argumentation framework $F = (A, R)$ with $A = \{1, 2, ..., n\}$, then the admissible set $S = \{i_{1}, i_{2}, ..., i_{k}\} \subset A$ is a complete extension in $F$ iff the following conditions hold:

(1) the column vector of $s$-block $M^{s} = M^{i_{1}, i_{2}, ..., i_{k}}_{j_{1}, j_{2}, ..., j_{h}}$ of $S$ corresponding to the non-zero
row vector of the $c$-block $M^{c} = M^{j_{1}, j_{2}, ..., j_{h}}_{j_{1}, j_{2}, ..., j_{h}}$ of $S$ is zero,

(2) the column vector of $s$-block $M^{s} = M^{i_{1}, i_{2}, ..., i_{k}}_{j_{1}, j_{2}, ..., j_{h}}$ of $S$ corresponding to the zero column
vector of the $c$-block $M^{c} = M^{j_{1}, j_{2}, ..., j_{h}}_{j_{1}, j_{2}, ..., j_{h}}$ of $S$ is non-zero,

\hspace{-10mm} where $A \setminus S = \{j_{1}, j_{2}, ..., j_{h}\}$.

For any p-admissible set $S$, the argument $i \in S$ is obviously acceptable $w.r.t$ $S$ and without indirect conflict with any argument of $S$. By definition 4,  the following lemma is an immediate result for $S$ to be a complete p-extension.

\vspace{2mm}
\hspace{-10mm} \textbf{Lemma 31} Let $F = (A, R)$ be an argumentation framework with $A = \{1, 2, ..., n\}$, $S = \{i_{1}, i_{2}, ..., i_{k}\} \subset A$, and $A \setminus S = \{j_{1}, j_{2}, ..., j_{h}\}$, then $S$ is a complete p-extension of $F$ iff $S$ is a p-admissible set and for each $1 \leq t \leq h$, $j_{t} \in S$ is not defended by $S$ or has indirect conflict with some element of $S$.

From theorem 14 and theorem 15, the argument $i$ indirectly defends the argument $j$ in $F$ if and only if $m(2)_{i, j} \neq 0$ in the matrix $M^{2}(F)$, and the argument $i$ indirectly attacks the argument $j$ in $F$ if and only if $m(O)_{i, j} \neq 0$ in the matrix $M^{O}(F)$. For the purpose of determining the complete $p$-extensions of an argumentation framework $F$, We introduce one class of blocks in the matrix $M^{2}(F)$ and two classes of blocks in the matrix $M^{O}(F)$.

\vspace{2mm}
\hspace{-10mm} \textbf{Definition 32} Let $F = (A, R)$ be an argumentation framework with $A = \{1, 2, ..., n\}$, $S = \{i_{1}, i_{2}, ..., i_{k}\} \subset A$, and $A \setminus S = \{j_{1}, j_{2}, ..., j_{h}\}$. Then, we say

\[
  M^{pcd}(2)^{i_{1}, i_{2}, ..., i_{k}}_{j_{1}, j_{2}, ..., j_{h}} = \left(\begin{array}{cccccc}
  m(2)_{i_{1}, j_{1}}&m(2)_{i_{1}, j_{2}}&.&.&.&m(2)_{i_{1}, j_{h}}\\
  m(2)_{i_{2}, j_{1}}&m(2)_{i_{2}, j_{2}}&.&.&.&m(2)_{i_{2}, j_{h}}\\
  .&.&.&.&.&.\\
  m(2)_{i_{k}, j_{1}}&m(2)_{i_{k}, j_{2}}&.&.&.&m(2)_{i_{k}, j_{h}}
  \end{array}\right)
\]
in the matrix $M^{2}(F)$ of $F$ is the $pcd$-block of $S$. And we say that the $k \times h$ block

\[
  M^{pca1}(O)^{i_{1}, i_{2}, ..., i_{k}}_{j_{1}, j_{2}, ..., j_{h}} = \left(\begin{array}{cccccc}
  m(O)_{i_{1}, j_{1}}&m(O)_{i_{1}, j_{2}}&.&.&.&m(O)_{i_{1}, j_{h}}\\
  m(O)_{i_{2}, j_{1}}&m(O)_{i_{2}, j_{2}}&.&.&.&m(O)_{i_{2}, j_{h}}\\
  .&.&.&.&.&.\\
  m(O)_{i_{k}, j_{1}}&m(O)_{i_{k}, j_{2}}&.&.&.&m(O)_{i_{k}, j_{h}}
  \end{array}\right)
\]
in the matrix $M^{O}(F)$ of $F$ is the $pca1$-block of $S$, the $h \times k$  block

\[
  M^{pca2}(O)^{j_{1}, j_{2}, ..., j_{k}}_{i_{1}, i_{2}, ..., i_{h}} = \left(\begin{array}{cccccc}
  m(O)_{j_{1}, i_{1}}&m(O)_{j_{1}, i_{2}}&.&.&.&m(O)_{j_{1}, i_{k}}\\
  m(O)_{j_{2}, i_{1}}&m(O)_{j_{2}, i_{2}}&.&.&.&m(O)_{j_{2}, i_{k}}\\
  .&.&.&.&.&.\\
  m(O)_{j_{h}, i_{1}}&m(O)_{j_{h}, i_{2}}&.&.&.&m(O)_{j_{h}, i_{k}}
  \end{array}\right)
\]
in the matrix $M^{O}(F)$ of $F$ is the $pca2$-block of $S$.

Namely, the elements appearing at the intersection of rows $i_{1}, i_{2}, ..., i_{k}$ and columns $j_{1}, j_{2}, ..., j_{h}$ in the matrix $M^{2}(F)$ form the $pcd$-block $M^{pcd}(2)^{i_{1}, i_{2}, ..., i_{k}}_{j_{1}, j_{2}, ..., j_{k}}$ of $S$, the elements appearing at the intersection of rows $i_{1}, i_{2}, ..., i_{k}$ and columns $j_{1}, j_{2}, ..., j_{h}$ in the matrix $M^{O}(F)$ form the $pca1$-block $M^{pca1}(O)^{i_{1}, i_{2}, ..., i_{k}}_{j_{1}, j_{2}, ..., j_{k}}$ of $S$, and the elements appearing at the intersection of rows $j_{1}, j_{2}, ..., j_{h}$ and columns $i_{1}, i_{2}, ..., i_{k}$ in the matrix $M^{O}(F)$ form the $pca2$-block $M^{pca2}(O)^{j_{1}, j_{2}, ..., j_{h}}_{i_{1}, i_{2}, ..., i_{k}}$ of $S$.

For convenience, we denote $M^{pcd}(2) = M^{pcd}(2)^{i_{1}, i_{2}, ..., i_{k}}_{j_{1}, j_{2}, ..., j_{k}}$, $M^{pca1}(O) = M^{pca1}(O)^{i_{1}, i_{2}, ..., i_{k}}_{j_{1}, j_{2}, ..., j_{k}}$ and $M^{pca2}(O) = M^{pca2}(O)^{j_{1}, j_{2}, ..., j_{h}}_{i_{1}, i_{2}, ..., i_{k}}$. In light the above lemma, we give the matric method of determining the p-complete extensions in an argumentation framework.

\vspace{2mm}
\hspace{-10mm} \textbf{Theorem 33} Let $F = (A, R)$ with $A = \{1, 2, ..., n\}$ be an argumentation framework, $S = \{i_{1}, i_{2}, ..., i_{k}\} \subset A$, and $A \setminus S = \{j_{1}, j_{2}, ..., j_{h}\}$, then a $p$-admissible set $S = \{i_{1}, i_{2}, ..., i_{k}\} \subset A$ is a p-complete extension in $F$ iff for each $j_{t} \in A \setminus S (1 \leq t \leq h)$, one of the following conditions hold:

(1) The column vector $M^{pcd}(2)_{*, j_{t}}$ of the $pcd$-block $M^{pcd}(2)$ in the matrix $M^{2}(F)$ is zero,

(2) The column vector $M^{pca1}(O)_{*, j_{t}}$ of the $pcd1$-block $M^{pca1}(O)$ in the matrix $M^{O}(F)$ is nonzero,

(3) The row vector $M^{pca2}(O)_{j_{t}, *}$ of the $pcd2$-block $M^{pca2}(O)$ in the matrix $M^{O}(F)$ is nonzero.

\vspace{2mm}
\hspace{-10mm} \textbf{Proof} Condition (1) ensures that $j_{t} \in A \setminus S$ is not defended by $S$, i.e., $j_{t}$ is not acceptable $w.r.t$ $S$, condition (2) means that $j_{t} \in S$ is attacked by some $i_{r} \in S (1 \leq r \leq k)$, and condition (3) indicates that $j_{t} \in S$ attacks some $i_{r} \in S (1 \leq r \leq k)$. Any one of these conditions implies that $j_{t}$ does not belong to $S$, and thus the p-admissible set $S$ is a complete p-extension.

Conversely, suppose that $S$ is a complete p-extension. Then, by lemma 31 $S$ is certainly a p-admissible set. Furthermore, any $j_{t} \in A \setminus S$ should not be defended by $S$ or has indirect conflict with some element of $S$ in light of definition 4. Therefore, there must be one of the three conditions to hold. \\

\hspace{-10mm} {\bf \Large 9. Conclusions and perspectives}

\vspace{5mm} In this paper, we introduced the matrix $M(F)$ of an argumentation framework $F = (A, R)$. By the matrices $M^{E}(F) = (m(E)_{s, t}) = M^{2}(F) + M^{4}(F) + ..., M^{2k}(F)$ and $M^{O}(F) = (m(O)_{s, t}) = M^{3}(F) + M^{5}(F) + ..., M^{2k - 1}(F)$, we give a matric method to find out the "controversial arguments". If we want to verify the indirect attack relation or indirect defence relation between two arguments, we only need to computer the related matrices $M^{O}(F)$, $M^{E}(F)$ and $M^{2}(F)$. In order to study the prudent extensions in an argumentation frameworks, we also define the $p$-block $M_{p}(O)^{i_{1}, i_{2}, ..., i_{k}}_{i_{1}, i_{2}, ..., i_{k}}$ of $S$, the $pad$-block $M^{pcd}(2) = M(2)^{i_{1}, i_{2}, ..., i_{k}}_{j_{1}, j_{2}, ..., j_{k}}$ of $S$, the $pca1$-block $M^{pca1}(O) = M^{pca1}(O)^{i_{1}, i_{2}, ..., i_{k}}_{j_{1}, j_{2}, ..., j_{k}}$ of $S$, and the $pca2$-block $M^{pca2}(O) = M^{pca2}(O)^{j_{1}, j_{2}, ..., j_{h}}_{i_{1}, i_{2}, ..., i_{k}}$ of $S$, and presented several theorems to decide various prudent extensions (stable p-extension, p-admissible set, preferred p-extension, complete p-extension) of the AF, by blocks of the matrices $M^{2}(F)$ and $M^{O}(F)$  of the AF and relations between these blocks. Comparing with traditional way to deal with argumentation frameworks such as reasoning and directed graph, our matric method has the advantage of computability. If we want to find out all the "controversial arguments" or prudent extensions, we only need to computer the related matrices and blocks. More important is that we can apply the matrix theory into the research of argumentation frameworks, and this may bring a mathematical period of argumentation frameworks.

Interestingly, the $p$-block $M(O)^{i_{1}, i_{2}, ..., i_{k}}_{i_{1}, i_{2}, ..., i_{k}}$ of $S$ corresponds to the determination for $S$ to be a stable p-extension (p-admissible set, complete p-extension respectively). And, the $pca1$-block $M(O)^{i_{1}, i_{2}, ..., i_{k}}_{j_{1}, j_{2}, ..., j_{k}}$ of $S$ is exactly the complementary block of the $pca2$-block $M(O)^{j_{1}, j_{2}, ..., j_{h}}_{i_{1}, i_{2}, ..., i_{k}}$ of $S$. Also, we can determine the complete p-extensions of an AF by computing and checking the $pad$-block $M^{pcd}(2)$, the $pca1$-block $M^{pca1}(O)$ and the $pca2$-block $M^{pca2}(O)$. These facts indicate that there is indeed a corresponding relation between the argumentation framework and its matrix. So, we can investigate the structure and properties of an argumentation framework by using the theory and method of matrix.

The prospective is that, we can find out the internal pattern of AFs and the relations between different objects which we concerned in AFs, by studying the related matrices and blocks of AFs. Our future goal is to develop the matrix method in the related areas of argumentation frameworks, such as argument acceptability, dialogue games, algorithm and complexity and so on [14, 19, 13, 12, 15, 16]. \\


\hspace{-8mm}{\large \bf  References}
\hspace{5mm}


\end{document}